\documentclass[10pt,twocolumn,letterpaper]{article}

\usepackage{cvpr}              
\definecolor{cvprblue}{rgb}{0.21,0.49,0.74}
\usepackage[pagebackref,breaklinks,colorlinks,allcolors=cvprblue]{hyperref}
\usepackage{pifont}
\newcommand{\cmark}{\ding{51}}  
\newcommand{\xmark}{\ding{55}}  
\usepackage{booktabs}
\usepackage{multirow}

\def\paperID{*****} 
\def\confName{CVPR}
\def\confYear{2026}

\title{UR-Bench: A Benchmark for Multi-Hop Reasoning over Ultra–High-Resolution Images}

\author{%
  \textbf{Siqi~Li$^{1,2}$}
  \quad
  \textbf{Xinyu~Cai$^{2, \dag}$}
  \quad
  \textbf{Jianbiao~Mei$^{1,2}$}
  \quad
  \textbf{Nianchen~Deng$^{2}$ }
  \quad
  \textbf{Pinlong~Cai$^{2}$ }\\
  \quad
  \textbf{Licheng~Wen$^{2}$ }
  \quad
  \textbf{Yufan~Shen$^{2}$ }
  \quad
  \textbf{Xuemeng~Yang$^{2}$ }
  \quad
  \textbf{Botian~Shi$^{2,*}$}
  \quad
  \textbf{Yong Liu$^{1,\dag}$}
  \\
  $^1$ Zhejiang University
  $^2$ Shanghai Artificial Intelligence Laboratory \\
}

\begin{document}
\maketitle

\renewcommand{\thefootnote}{\relax}
\footnotetext{* project leader, $\dag$ corresponding author}
\renewcommand{\thefootnote}{\arabic{footnote}}

\begin{abstract}


Recent multimodal large language models (MLLMs) show strong capabilities in visual-language reasoning, yet their performance on ultra–high-resolution imagery remains largely unexplored. 
Existing visual question answering (VQA) benchmarks typically rely on medium-resolution data, offering limited visual complexity. 
To bridge this gap, we introduce \textbf{Ultra-high-resolution Reasoning Benchmark (UR-Bench)}, a benchmark designed to evaluate the reasoning capabilities of MLLMs under extreme visual information. 
UR-Bench comprises two major categories—\textit{Humanistic Scenes} and \textit{Natural Scenes}—covering four subsets of ultra–high-resolution images with distinct spatial structures and data sources. 
Each subset contains images ranging from hundreds of megapixels to gigapixels, accompanied by questions organized into three levels, enabling evaluation of models’ reasoning capabilities in ultra–high-resolution scenarios.
We further propose an \textbf{agent-based framework} in which a language model performs reasoning by invoking external visual tools. In addition, we introduce Semantic Abstraction and Retrieval tools that enable more efficient processing of ultra–high-resolution images.
We evaluate state-of-the-art models using both an end-to-end MLLMs and our agent-based framework, demonstrating the effectiveness of our framework.

\end{abstract}
    
\section{Introduction}









Recent advances in multimodal large language models (MLLMs) have demonstrated remarkable performance across a wide range of multimodal tasks, including visual question answering (VQA) and reasoning \cite{xu2025qwen3, comanici2025gemini, wang2025internvl3}. 
However, despite these successes, there is a growing interest in developing \textit{``think-with-image''} models \cite{su2025openthinkimg, ma2024taco, su2025thinking, li2025imagine, gupta2023visual}, which aim to enhance the model's ability to reason explicitly over visual content. 
Specifically, these methods often adopt an agent-style paradigm, wherein the model serves as a central controller that can invoke external tools as needed, thereby enabling more flexible and modular multimodal reasoning.

Despite these advances, existing VQA benchmarks remain limited in visual complexity \cite{STVQA, DocVQA, OCRVQA, TextVQA}. 

Although some datasets, such as V*Bench \cite{vstar} and HR-Bench \cite{HRbench}, claim to include high-resolution images, they typically rely on medium-resolution data whose spatial scale and scene density remain far below those of real-world ultra–high-resolution imagery.
To comprehensively evaluate the capabilities of MLLMs and the “think-with-image” methods, and push the boundaries of multimodal reasoning
we propose the \textbf{Ultra-high-resolution Reasoning Benchmark (UR-Bench)}, a benchmark tailored for multi-hop reasoning over ultra-high-resolution images.

\begin{figure*}[t]
    \centering
    \includegraphics[width=1\linewidth]{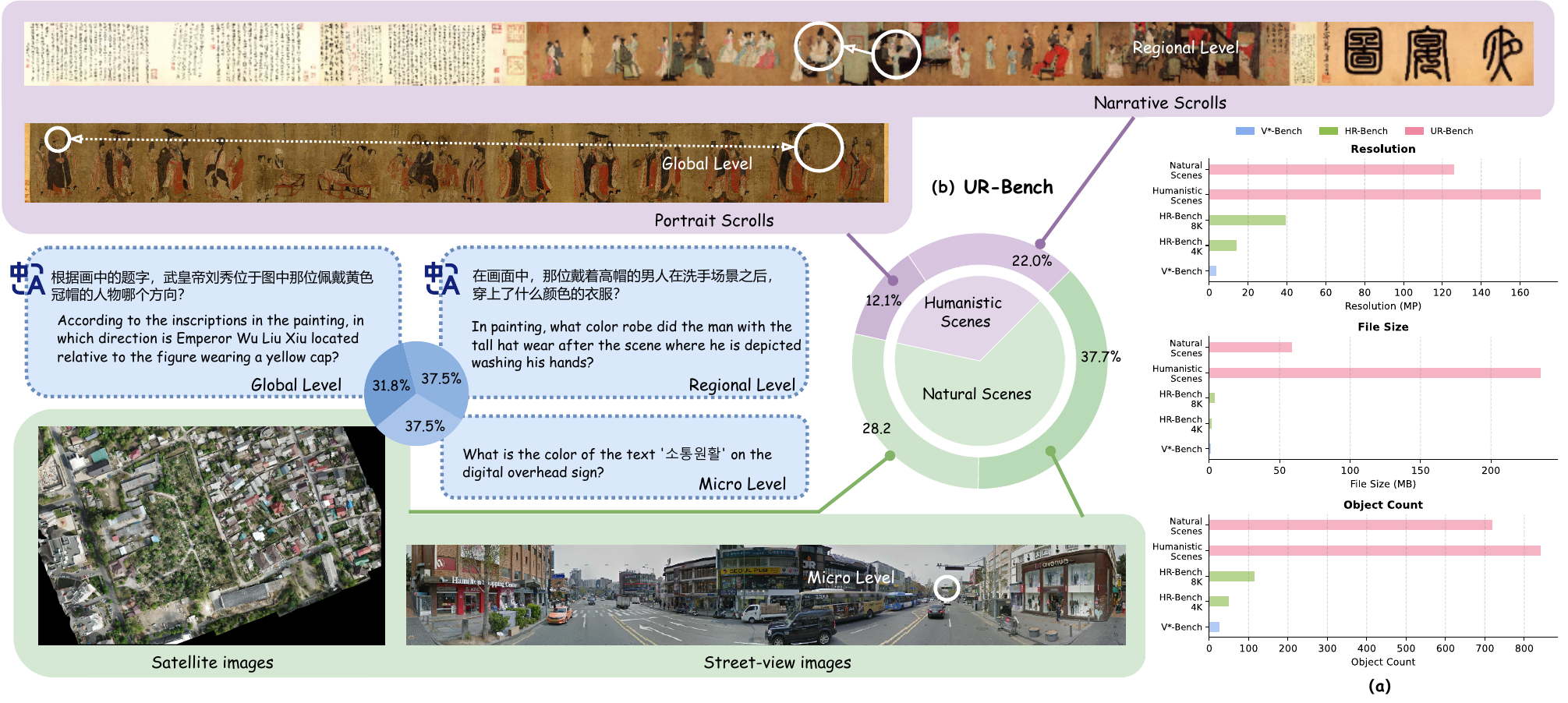}
    \caption{(a) Mean resolution (MP), file size (MB), and object count of high-resolution image benchmarks. (b) Two categories and four subsets of UR-Bench, with tasks organized across three difficulty levels.}\label{fig:highres_comparison}
\end{figure*}

\textbf{UR-Bench} is composed of two major categories and four subsets of ultra-high-resolution images, each defined by distinct spatial structures and data sources, as shown in Fig.\ref{fig:highres_comparison}. The \textbf{Humanistic Scenes} category features large-scale ancient Chinese scroll paintings, which include both \textit{Narrative Scrolls} and \textit{Portrait Scrolls}. The \textbf{Natural Scenes} category encompasses extensive landscape imagery from \textit{Satellite Images} and \textit{Street-view Images}. Across all subsets, the images are exceptionally large—often reaching hundreds of megapixels or hundreds of megabytes—and exhibit rich visual and spatial details, posing substantial challenges for visual understanding and reasoning.
To further differentiate reasoning complexity, questions are categorized into three levels, corresponding to increasing demands on spatial integration and multi-step inference.

UR-Bench poses significant challenges for end-to-end MLLMs. Directly processing such enormous images often exceeds the token capacity of most models, while simple compression or downsampling leads to information loss and degraded reasoning performance. Moreover, existing agent-based frameworks, such as OpenThinkIMG \cite{su2025openthinkimg}, struggle to handle images with such ultra–high resolutions.
To overcome these limitations, we propose an \textbf{agent-based framework} for multi-hop reasoning over ultra-high-resolution images.
Within this framework, the agent itself is a large language model (LLM) that interacts with the multimodal environment purely through language. 
It autonomously plans the sequence of operations and invokes specialized visual tools to perform perception and reasoning tasks in a modular, flexible manner. 
Unlike existing agent frameworks, our approach incorporates a Semantic Abstraction and Retrieval Tool, which maps complex and large-scale visual information into the language space during reasoning. 
This enables more effective processing of ultra–high-resolution input images and facilitates the solution of multi-hop reasoning tasks.
An illustrative example of this process is shown in Figure~\ref{fig:pipeline}.

Our main contributions are summarized as follows:
\begin{itemize}
    \item We introduce UR-Bench, a \textbf{benchmark for ultra–high-resolution multi-hop reasoning}, where individual image files range from several megabytes to over 1 GB and exhibit high information density. The benchmark incorporates three levels of reasoning complexity, enabling fine-grained evaluation under extreme visual conditions.  
    \item We propose an \textbf{automated data engine} for generating multi-hop reasoning questions over ultra–high-resolution images, capable of automatically producing questions with varying levels of reasoning difficulty.
    \item We propose an \textbf{agent-based framework} that enables LLMs to autonomously plan and coordinate tool-based operations. 
    The framework emphasizes semantic decomposition of ultra-large-scale visual information through the Semantic Abstraction and Retrieval Tool, enabling efficient perception and reasoning over ultra–high-resolution images.  
\end{itemize}
\section{Related Work}



\subsection{Visual Question Answering Benchmarks}
A wide range of Visual Question Answering (VQA) benchmarks have been developed to evaluate the multimodal reasoning capabilities of models across diverse visual and linguistic contexts. 
ChartQA~\cite{ChartQA} and related datasets assess the understanding of structured information in charts and plots.
TextVQA~\cite{TextVQA} emphasizes reading and reasoning over scene text. 
DocVQA~\cite{DocVQA} extends this paradigm to document images involving layout-aware and multi-hop reasoning. 
CharXiv~\cite{wang2024charxiv} explores scholarly figures and multi-page documents, testing fine-grained cross-modal reasoning. 
However, they rarely probe the visual limits of MLLMs, as their images remain small, simple, and low in informational density, offering limited insight into model performance in complex, information-rich scenes.
To address these limitations, a new generation of high-resolution (HR) and specialized benchmarks has been introduced, aiming to better evaluate fine-grained perception and reasoning—examples include V*Bench~\cite{vstar} and HR-Bench~\cite{HRbench}. 
Nevertheless, the spatial scale and scene complexity of these benchmarks remain far below those encountered in real-world ultra–high-resolution scenarios.

\subsection{Think-with-image}
Recent advances in multimodal reasoning have emphasized ``think-with-image'', where reasoning is explicitly grounded in visual representations.
Notable methods include OpenThinkIMG \cite{su2025openthinkimg}, which leverages visual tool reinforcement learning for image-grounded reasoning; Imagine \cite{li2025imagine}, which visualizes intermediate reasoning steps in spatial layouts; TACO \cite{ma2024taco}, which learns multimodal action models via synthetic chains-of-thought-and-action; and VisProg \cite{gupta2023visual}, which integrates visual planning with symbolic reasoning.
These works collectively move multimodal models from passive perception toward active, image-grounded reasoning.
However, they primarily operate on standard-sized images and are unable to handle VQA tasks involving ultra-high-resolution images.

\section{Benchmark}




\subsection{Subsets and Data Sources}

UR-Bench consists of two major categories and four subsets, each featuring ultra-high-resolution images with distinct spatial characteristics and data sources, as shown in Fig.\ref{fig:highres_comparison}.
\begin{itemize}
    \item \textbf{Humanistic Scenes} 
    This subset consists of large-scale ancient Chinese scroll paintings collected from Wikipedia. The average resolution is 48,685 $\times$ 2,821 pixels ($\sim$171 MP), with the highest resolution reaching up to 718 MP. It includes two major categories: 
    \begin{itemize}
    \item\textbf{Narrative Scrolls} typically feature continuous storytelling or transitions across multiple spaces, depicting scenes such as urban life, court activities, and mythological stories. 
    \item\textbf{Portrait Scrolls} are typically composed of a series of independent figures (or animals), with each section depicting a distinct subject or motif, arranged sequentially as a whole.
    \end{itemize}
    
    \item \textbf{Natural Scenes} 
    This subset combines large-scale natural and urban landscape images. The images have an average size of 19,443 $\times$ 7,557 pixels ($\sim$125.95 MP), with the maximum file size reaching 432.29 MB.
    \begin{itemize}
    \item\textbf{Satellite images} are sourced from the SUNet dataset~\cite{shao2021sunet}, which contains diverse aerial views of urban areas and terrains.
    \item\textbf{Street-view images} obtained via the Google Street View Static API,  covering approximately 90 bustling streets worldwide and containing rich visual information. 
    \end{itemize}

\end{itemize}


\begin{table}[htbp]
\centering
\caption{A comparison of high-resolution image benchmarks.}
\resizebox{1\linewidth}{!}{
\begin{tabular}{@{}lccc@{}}
\toprule
Benchmark                & Multilingual    & Multi-source    & Difficulty Grading \\ \midrule
VSTAR-Bench \cite{vstar} & \xmark          & \xmark          & \xmark             \\
HR-Bench \cite{HRbench}  & \xmark          & \xmark          & \xmark             \\
\textbf{UR-Bench (Ours)} & \textbf{\cmark} & \textbf{\cmark} & \textbf{\cmark}    \\ \bottomrule
\end{tabular}}
\label{tab:compare_bench}
\end{table}


Across all subsets, UR-Bench demonstrates a consistently ultra-large scale, with individual image files ranging from several megabytes to over 1 GB. Such enormous resolutions pose significant challenges for patch-based reasoning and multi-region visual understanding. 
We compare the image sizes in our benchmark with those in existing high-resolution benchmarks, including V*Bench\cite{vstar} and HR-Bench\cite{HRbench} in Fig. \ref{fig:highres_comparison} (a). The images in our benchmark are significantly larger than those in the other benchmarks.

In addition to the visual scale, we further quantify the semantic richness of each dataset by introducing the \textit{Object Count} metric. 
Specifically, we employ the GroundingDINO~\cite{liu2024grounding} zero-shot object detection model to identify and localize all detectable entities within each image. 
The average and maximum \textit{Object Count} values reported in Fig.~\ref{fig:highres_comparison} (a) reflect the number of distinct object instances present per image, which serves as an indicator of semantic density and visual clutter. 
UR-Bench exhibits substantially higher object counts, indicating more complex and information-dense visual environments.

In addition to incorporating diverse data sources, UR-Bench also includes multilingual questions and categorizes them according to difficulty levels. Table \ref{tab:compare_bench} presents a comparison with other high-resolution benchmarks.

\subsection{Task Categorization}


Furthermore, we categorize all questions in the benchmark into three types based on the required reasoning process and spatial dependencies:

\begin{itemize}
    \item \textbf{Micro Level.} The answer can be derived from a single region in the image without requiring cross-region association or multi-step reasoning. Although no complex reasoning is needed, locating the correct region in ultra-high-resolution images remains challenging due to the vast spatial scale and sparsity of relevant content, making this type akin to a ``needle-in-a-haystack'' search.
    \item \textbf{Regional Level.} This type of question builds on Micro Level and requires 2–3 incremental visual-reasoning steps to arrive at an answer based on observations of \textbf{adjacent regions}. It is characterized by spatially contiguous visual cues that must be processed in a sequential manner.
    \item \textbf{Global Level.} The question involves integrating information from two or more distant, \textbf{non-adjacent regions} of the image. The reasoning often depends on spatial relationships or positional comparisons across regions, reflecting a higher level of reasoning complexity.
\end{itemize}

\begin{figure}[t]
    \centering
    \includegraphics[width=1\linewidth]{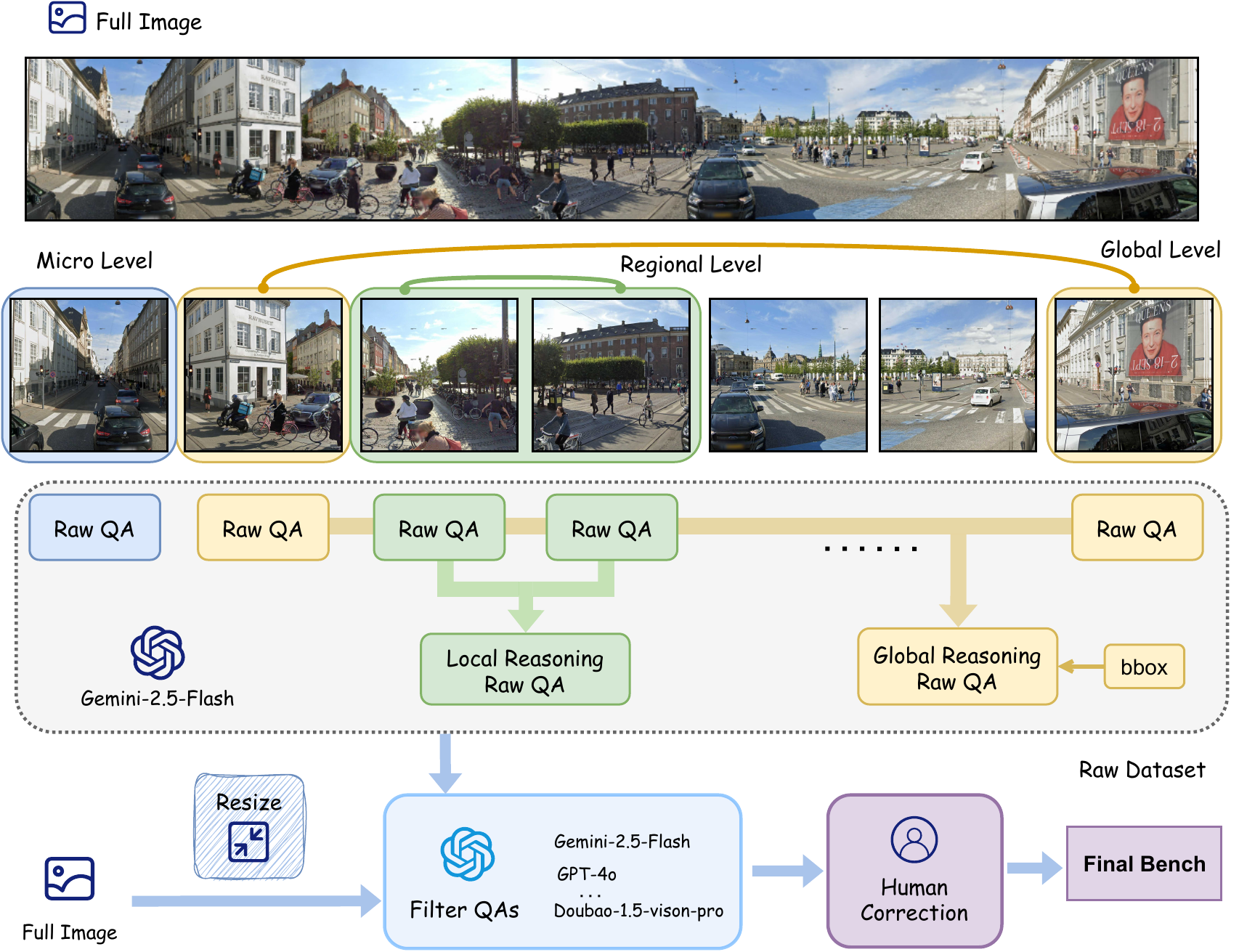}
    \caption{Overview of the data construction pipeline. }\label{fig:data_construction}
\end{figure}

\subsection{Data Engine}

UR-Bench is constructed through a hierarchical pipeline that integrates image partitioning, multimodal QA generation, reasoning-level composition, automatic filtering, and human calibration, as illustrated in Fig.~\ref{fig:data_construction}.

We first divide each ultra-high-resolution image into multiple non-overlapping tiles to enable effective processing within the input limits of vision-language models (VLMs). Each tile is then fed into {Gemini-2.5-Flash} \cite{comanici2025gemini}, which generates 0–2 question–answer (QA) pairs describing fine-grained visual details such as objects and attributes. 

For Regional Level and Global Level questions, to go beyond purely local perception, we construct higher-level reasoning questions by jointly inputting each image patch, its corresponding QA pair, and the patch’s bounding box within the original image into Gemini-2.5-Flash.
This design explicitly provides the model with both the visual content of each region and its spatial position in the overall scene, enabling the generation of local reasoning and global reasoning QAs that require integrating information across multiple regions or relating local evidence to global context. 

Subsequently, we automatically filter the generated raw dataset using MLLMs (e.g., GPT-4o) to remove overly simple or low-quality questions. At this stage, the compressed images (sized appropriately for end-to-end inference) and their corresponding QA pairs are jointly fed into the large model.
Finally, human experts manually verify and refine the remaining QA pairs, yielding the final benchmark dataset.

\section{Method}


\begin{figure*}[t]
    \centering
    \includegraphics[width=1\linewidth]{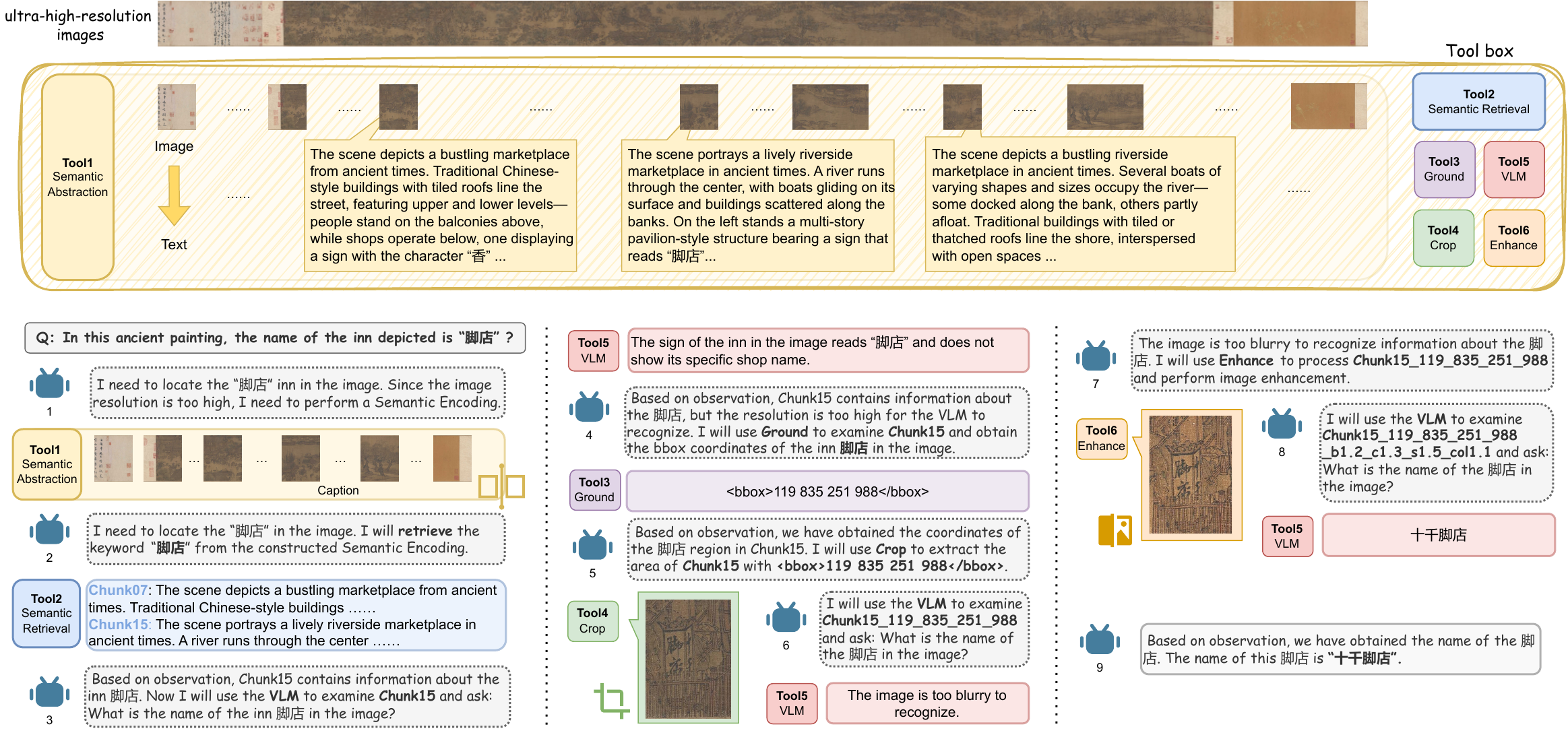}
    \caption{Illustration of the agent-based framework for ultra-high-resolution image QA. The agent operates through natural-language reasoning and dynamically invokes external visual tools to handle large-scale images.}\label{fig:pipeline}
\end{figure*}


The ultra–high-resolution images in UR-Bench pose substantial challenges to end-to-end reasoning, as most open-source and closed-source MLLMs have limited token capacities. Directly feeding such massive images into MLLMs often exceeds their input limits, whereas simple downsampling or compression inevitably causes severe information loss, undermining both perception and reasoning quality. Furthermore, existing agent-based frameworks, such as OpenThinkIMG~\cite{su2025openthinkimg}, are also ill-equipped to manage such data, since their decision models rely on VLMs and their available tools lack the capability to process images at this scale.

To overcome these limitations, we propose an agent-based framework designed for multi-hop reasoning over ultra–high-resolution images. Unlike other agent-based frameworks, our decision model is a purely language-based model that does not take images as direct input; instead, it progressively explores and interprets complex visual scenes by invoking a specialized set of tools. The process by which an agent utilizes tools and reasoning is illustrated in Fig \ref{fig:pipeline}.

Our toolset comprises two categories of tools: the Semantic Abstraction and Retrieval Tools and the Visual Tools, with their input and output formats summarized in Table \ref{tool}.

\begin{table}[t]
\centering\resizebox{1\linewidth}{!}{
\begin{tabular}{@{}lll@{}}
\hline
\textbf{Tool} & \textbf{Input} & \textbf{Output}  \\
\hline
\textbf{Semantic Abstraction} & image + chunk number & json  \\
\textbf{Semantic Retrieval} & json + query + topk & top-k retrieved texts  \\
\textbf{VLM} & image + prompt & text answer  \\
\textbf{Crop} & image + bbox coordinates & cropped image  \\
\textbf{Ground} & image + keyword & bbox coordinates  \\
\textbf{Enhance} & \begin{tabular}[c]{@{}l@{}}image + brightness + contrast \\  + sharpness + color\end{tabular} & enhanced image  \\
\hline
\end{tabular}}
\caption{Toolset Overview. This table presents our toolset, detailing their input requirements and output formats.}\label{tool}
\end{table}

\subsection{Semantic Abstraction and Retrieval.}

Existing visual retrieval-augmented generation (V-RAG) methods~\cite{yu2024visrag} face significant challenges when dealing with ultra-high-resolution images. 
Raw visual tokens cannot be directly used for text retrieval because they lack a compact, semantically meaningful representation suitable for similarity matching.
Directly embedding all pixels quickly exceeds model token limits, dense patch-level embeddings impose heavy computational and memory costs, and simple resizing or tiling often discards fine-grained semantic details. 

To address these limitations, we introduce a \textbf{Semantic Abstraction and Retrieval Tool}, which enables semantic-level exploration in our agent framework, converting visual information into a language-based representation that can be efficiently retrieved.

\subsubsection{Semantic Abstraction}

The Semantic Abstraction Tool processes an ultra-high-resolution input image $I$ by first dividing it into $N$ visual chunks $C=\{c_i\}$ determined by an LLM agent. 

The image is segmented into $N$ visual chunks $C=\{c_1, c_2, \dots, c_N\}$ such that:
\begin{equation}
I=\bigcup_{i=1}^{N} c_i
\end{equation}

Each chunk $c_i$ is transformed into a semantically rich caption ($L_i$) using the Qwen2.5-VL-7B model~\cite{bai2025qwen2}, mapping the visual data into the language space $L=\{L_1, L_2, \dots, L_N\}$. We denote the equivalent token counts \textbf{before} and \textbf{after} this transformation as $T_{\text{raw}}(I)$ and $T_{\text{lang}}(L)$, and the maximum model capacity as $T_{\max}$. 

Due to the size constraints of ultra-high-resolution imagery, the raw token requirement far exceeds model capacity ($T_{\text{raw}}(I) \gg T_{\max}$), making direct processing infeasible. The resulting language representation, however, is compact:
\begin{equation}
T_{\text{lang}}(L) \ll T_{\text{raw}}(I),\qquad T_{\text{lang}}(L_i)\le T_{\max}
\end{equation}

This process performs semantic abstraction, significantly reducing the representational scale while preserving essential semantic information, thereby facilitating subsequent text-based retrieval.
Through the Semantic Abstraction Tool, the agent obtains a JSON file containing a set of paired representations $(c_i, L_i)$, where $i = 1, \ldots, N$.

\subsubsection{Semantic Retrieval}

Given a textual query $Q$, the Semantic Retrieval Tool employs the BGE-M3 embedding model~\cite{chen2024bge} to perform semantic similarity matching across the set of pre-generated captions $L=\{L_1, L_2, \dots, L_N\}$. Both the query and the individual captions are first encoded into embedding vectors, $\mathbf{e}_Q$ and $\mathbf{e}_{L_i}$, respectively.

The semantic similarity between the query and a caption is quantified using the cosine similarity metric:
\begin{equation}
S(Q, L_i) = \frac{\mathbf{e}_Q \cdot \mathbf{e}_{L_i}}{\|\mathbf{e}_Q\|_2 \cdot \|\mathbf{e}_{L_i}\|_2}.
\end{equation}
The tool then identifies and selects the top-$k$ most relevant captions, denoted as $L_{\text{top-}k} \subset L$, based on the calculated similarity scores.

The visual regions corresponding to the top-ranked captions are aggregated to form the image region relevant to the query:
\begin{equation}
I_{\text{relevant}} = \bigcup_{L_i \in L_{\text{top-}k}} c_i,
\end{equation}
where $c_i$ is the visual chunk from which caption $L_i$ was generated. This effectively identifies the most semantically pertinent regions within the original ultra-high-resolution image.

Through this efficient process, the Semantic Retrieval Tool enables the agent to perform localized reasoning directly in the language space, facilitating scalable and efficient exploration of ultra-high-resolution imagery content.

\begin{table*}[t]
\centering\resizebox{1\linewidth}{!}{
\begin{tabular}{@{}lccccccc@{}}
\toprule
\textbf{Method}              & \textbf{Portrait Scrolls} & \textbf{Narrative Scrolls} & \textbf{Humanistic Scenes} & \textbf{Satellite images} & \textbf{Street-view images} & \textbf{Natural Scenes} & \textbf{Overall} \\ \midrule
\multicolumn{8}{c}{\textbf{agent framework (Ours)}}                                                                                                                                                                       \\ \midrule
gpt4o                        & 32.81                     & 27.59                      & 29.44                      & \textbf{39.72}            & 41.67                       & \textbf{40.84}          & 36.95            \\
doubao-seed-1-6-thinking     & 39.06                     & 31.90                      & 34.44                      &                           29.53& 44.27                       &                         37.95&                  36.75\\
Qwen3-235B-A22B-Instruct     & \textbf{46.88}            & \textbf{35.34}             & \textbf{39.44}             & 30.70                     & \textbf{47.40}              & 40.25                   & \textbf{39.97}   \\
gemini-2.5-flash-thinking    & 31.25                     & 25.00                      & 27.22                      &                           25.53& 27.08                       &                         26.14&                  26.69\\
gpt-4.1 (2025-04-14)         & 43.75                     & 33.62                      & 37.22                      & 34.75                     & 41.67                       & 38.71                   & 38.20            \\
claude-sonnet-4 (2025-05-14) & 37.50                     & 31.03                      & 33.33                      & 34.04                     & 44.79                       & 40.19                   & 37.85            \\
DeepSeek-R1                  & 35.94                     & 34.48                      & 35.00                      &                           31.25& 40.62                       &                         36.60&                  36.06\\ \midrule
\multicolumn{8}{c}{\textbf{end-to-end}}                                                                                                                                                                                   \\ \midrule
gpt4o                        & 12.50                     & 16.52                      & 15.08                      & 32.88                     & 18.59                       & 24.71                   & 21.43            \\
o3                           & 12.50                     & 24.35                      & 20.11                      & 15.28                     & 21.32                       & 18.73                   & 19.20            \\
doubao-seed-1-5-vison-pro    & 23.44                     & 22.61                      & 22.91                      & 34.69                     & 23.98                       & 28.57                   & 26.64            \\
grok-2-vision-1212           & 26.56                     & 19.83                      & 22.22                      & 24.49                     & 15.58                       & 19.39                   & 20.35            \\
qwen-2.5-vl-72b              & 20.31                     & 21.74                      & 21.23                      & 33.33                     & 32.14                       & 32.65                   & 28.76            \\
qwen-2.5-vl-32b              & 26.56                     & 20.69                      & 22.78                      & 31.97                     & 26.13                       & 28.63                   & 26.64            \\
gemini-2.5-flash-thinking    & 21.88                     & 20.00                      & 20.11                      & 15.65                     & 24.62                       & 20.78                   & 20.55            \\
claude-sonnet-4 (2025-05-14) & 34.38                     & 30.17                      & 31.66                      & 31.97                     & 14.07                       & 21.73                   & 25.12            \\ 
\bottomrule
\end{tabular}}
\caption{Accuracy scores on the UR-Bench. The results encompass end-to-end evaluations of both closed-source and open-source MLLMs, as well as the performance of our proposed agent framework equipped with different decision models.}\label{acc}
\end{table*}

\subsection{Visual Tool}

In addition, our toolset also includes the following basic Visual Tools for performing fundamental image localization, editing, and question-answering tasks.



\textbf{VLM.}
A visual–language model built upon Qwen2.5-VL-7B~\cite{bai2025qwen2}, employed to perform visual question answering for localized image patches $I$ given a query $q$.

\textbf{Ground.}
The tool enables integrated language and visual perception through text-driven object detection. Leveraging the GroundingDINO model~\cite{liu2024grounding}, it identifies visual regions that correspond to the given textual input. For an input image $I$ and text query $q$, the tool detects instances of the described objects and outputs their corresponding bounding boxes $b$.

\textbf{Crop.}
This tool performs image cropping based on provided bounding boxes, enabling the agent to isolate and analyze specific regions of interest. Given an input image $I$ and the corresponding bounding boxes $b$ (defining the target region), it extracts the designated rectangular subregion and outputs the cropped images $I_{{crop}}$.

\textbf{Enhance.}
This tool applies image enhancement operations—covering brightness, contrast, sharpness, and color adjustments—to mitigate visual degradation while preserving a natural appearance. Given an input image $I$ and enhancement parameters $b$, $c$, $s$, and $col$ (corresponding to brightness, contrast, sharpness, and color, respectively), the tool processes the image and produces an enhanced output $I_{\text{enhance}}$.

Additional implementation details and tool-specific hyperparameters are provided in the Appendix for completeness and reproducibility.

\section{Experiment}







\subsection{Experiment Setup}

We evaluate our benchmark across a broad range of both open-source and closed-source MLLMs. Specifically, we conduct end-to-end testing on the following models: GPT-4o \cite{hurst2024gpt}, o3 \cite{openai2025o3}, Doubao-Seed-1.5-Vision-Pro, Grok-2-Vision, Qwen-2.5-VL-72B \cite{bai2025qwen2}, Qwen-2.5-VL-32B \cite{bai2025qwen2}, Gemini-2.5-Flash-Thinking \cite{comanici2025gemini}, and Claude-Sonnet-4 (2025-05-14) \cite{anthropic2025claude37}.  
Given that our benchmark contains ultra-high-resolution images that may exceed the token limits of these models, we resize each image proportionally to ensure its visual token count matches the maximum visual token capacity of each model during inference. 

We further assess our proposed agent framework, where the reasoning and decision-making processes are handled by pure language models. The decision models include GPT-4o \cite{hurst2024gpt}, Doubao-Seed-1.6-Thinking, Qwen3-235B-A22B-Instruct, GPT-4.1 (2025-04-14), Claude-Sonnet-4 (2025-05-14), Gemini-2.5-Flash-Thinking \cite{comanici2025gemini}, and DeepSeek-R1 \cite{guo2025deepseek}.  
In this setting, the agent framework processes the visual environment through modular perception tools and delegates high-level reasoning and decision-making to the language model. 
All inferences are conducted through official APIs under consistent configurations.
All questions in our benchmark are 4-option single-choice questions, and the final evaluation metric is accuracy, defined as the proportion of correctly answered questions among all valid responses.

\subsection{Overall Results}

We present the main results on the UR-Bench in Table~\ref{acc}. Our analysis focuses on three key observations.

\paragraph{End-to-End MLLMs Show Limited Capability}
As evidenced by the ``end-to-end'' results in Table~\ref{acc}, {the performance of current MLLMs still has significant room for improvement} when faced with ultra-high-resolution images. Despite being powerful models, methods like {gpt4o} and {claude-sonnet-4} achieve low overall accuracy scores of 21.43 and 25.12, respectively. This poor performance highlights the inherent challenge these models face in processing gigapixel-level images in a single pass, likely leading to severe information loss and an inability to perceive fine-grained details.

\paragraph{Our Agent Framework Achieves Superior Performance}
In stark contrast, {our proposed agent framework is better} and demonstrates a substantial performance leap across all categories. By systematically decomposing the complex task and strategically processing the high-resolution input, our framework consistently outperforms all end-to-end baselines. For instance, using {gpt4o} as the decision model within our framework yields an overall score of 36.95, a +15.52 percentage point improvement over its end-to-end counterpart. Our framework equipped with {Qwen3-235B-A22B-Instruct} achieves the highest {Overall} score of {39.97}, establishing a new state-of-the-art on this challenging benchmark.

\paragraph{Analysis of Subset Performance}
The granular results across the different subsets reveal further insights into the models' capabilities. The \textbf{Humanistic Scenes} subset, characterized by the intricate, fine-grained details of ancient Chinese scroll paintings, proved particularly challenging. Our agent framework excelled in this domain, especially when equipped with {Qwen3-235B-A22B-Instruct}, which achieved the top scores on {Portrait Scrolls} (46.88), {Narrative Scrolls} (35.34), and the aggregate category (39.44). This suggests our framework is highly effective at parsing the complex narratives and detailed figures inherent in these artistic works. Conversely, in the \textbf{Natural Scenes} subset, which combines large-scale aerial and ground-level imagery, {gpt4o} within our framework showed a particular strength, achieving the highest score for this category (40.84), driven by strong performance on both {Satellite images} (39.72) and {Street-view images} (41.67). This demonstrates our framework's flexibility in leveraging different models' strengths for varied visual contexts, from art to geospatial analysis.

\subsection{Analysis by Reasoning Complexity}

\begin{table}[t]
\centering
\centering\resizebox{1\linewidth}{!}{
\begin{tabular}{@{}lccc@{}}
\toprule
\textbf{Method}              & \textbf{Micro Level} & \textbf{Regional Level} & \textbf{Global Level} \\ \midrule
\multicolumn{4}{c}{\textbf{agent framework (Ours)}}                                                   \\ \midrule
gpt4o                        & 41.07                & 33.48                   & 38.47                 \\
doubao-seed-1-6-thinking     &                      \textbf{48.19}&                         29.08&                       34.44\\
Qwen3-235B-A22B-Instruct     & 45.59& 37.48                   & 35.21                 \\
gemini-2.5-flash-thinking    &                      28.11&                         20.91&                       31.99\\
gpt-4.1 (2025-04-14)         & 41.27                & 35.89                   & \textbf{38.81}        \\
claude-sonnet-4 (2025-05-14) & 39.97                & \textbf{37.80}          & 36.07                 \\
DeepSeek-R1                  &                      38.06&                         31.86&                       36.64\\ \midrule
\multicolumn{4}{c}{\textbf{end-to-end}}                                                               \\ \midrule
gpt4o                        & 17.05                & 21.80                   & 28.90                 \\
o3                           & 17.32                & 15.13                   & 16.23                 \\
doubao-seed-1-5-vison-pro    & 25.14                & 21.32                   & 33.63                 \\
grok-2-vision-1212           & 13.39                & 17.56                   & 25.21                 \\
qwen-2.5-vl-72b              & 24.88                & 26.81                   & 33.76                 \\
qwen-2.5-vl-32b              & 23.20                & 24.71                   & 29.67                 \\
gemini-2.5-flash-thinking    & 20.80                & 19.26                   & 17.02                 \\
claude-sonnet-4 (2025-05-14) & 21.32                & 19.71                   & 30.52                 \\ \bottomrule
\end{tabular}}
\caption{Model performance comparison across Micro Level, Regional Level and Global Level.}\label{tab:difficulty}
\end{table}

Table~\ref{tab:difficulty} presents the updated model performance across Micro, Regional, and Global levels. Overall, our agent framework continues to outperform end-to-end models across all reasoning complexities, highlighting the benefits of explicit region retrieval and structured multi-step reasoning. For instance, gpt4o improves from 17.05/21.80/28.90 in the end-to-end setting to 41.07/33.48/38.47 with the agent, while doubao-seed-1-6-thinking reaches 48.19 on the Regional Level, indicating strong multi-step reasoning capabilities. Within the agent framework, different models show complementary strengths: Qwen3-235B-A22B-Instruct achieves the highest Micro-Level accuracy (45.59), demonstrating precise fine-grained perception; claude-sonnet-4 attains the best Regional-Level accuracy (37.80), reflecting effective short-horizon reasoning over adjacent regions; and gpt-4.1 performs best on the Global Level (38.81), suggesting superior integration across distant regions. Other agent models such as DeepSeek-R1 and gemini-2.5-flash-thinking also show competitive performance, illustrating that multiple models can contribute to multi-level reasoning. In contrast, end-to-end models remain significantly lower across all levels, rarely exceeding 34\%, emphasizing their difficulty in locating sparse regions and performing cross-region reasoning without agent guidance. Interestingly, some end-to-end models achieve relatively higher scores on Global-Level questions (e.g., qwen-2.5-vl-72b: 33.76), likely because reasoning over coarse spatial relations is less dependent on precise region localization. These results reinforce that ultra-high-resolution VQA benefits greatly from agent-based decomposition and multi-level evaluation is crucial to capture the diverse strengths of modern MLLMs.

\begin{figure*}[t]
    \centering
    \includegraphics[width=1\linewidth]{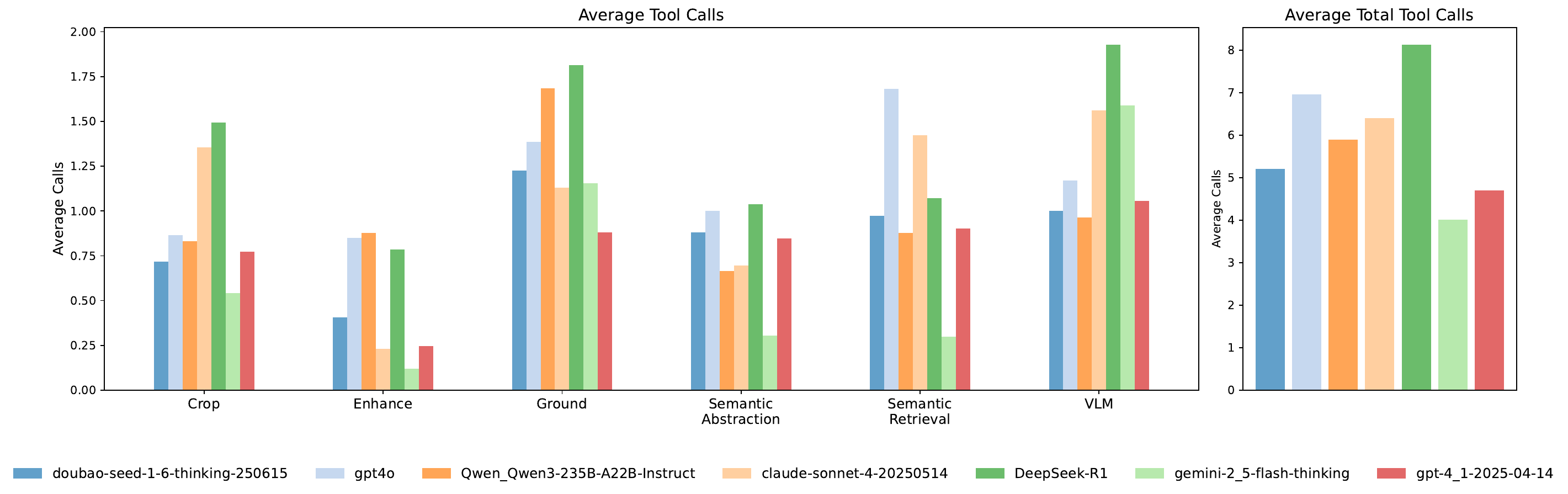}
    \caption{Comparison of tool invocation frequencies under the Street-view Images subset, using different models as the decision model within our agent framework.}\label{fig:tool_usage}
\end{figure*}
\begin{figure*}[h]
    \centering
    \includegraphics[width=1\linewidth]{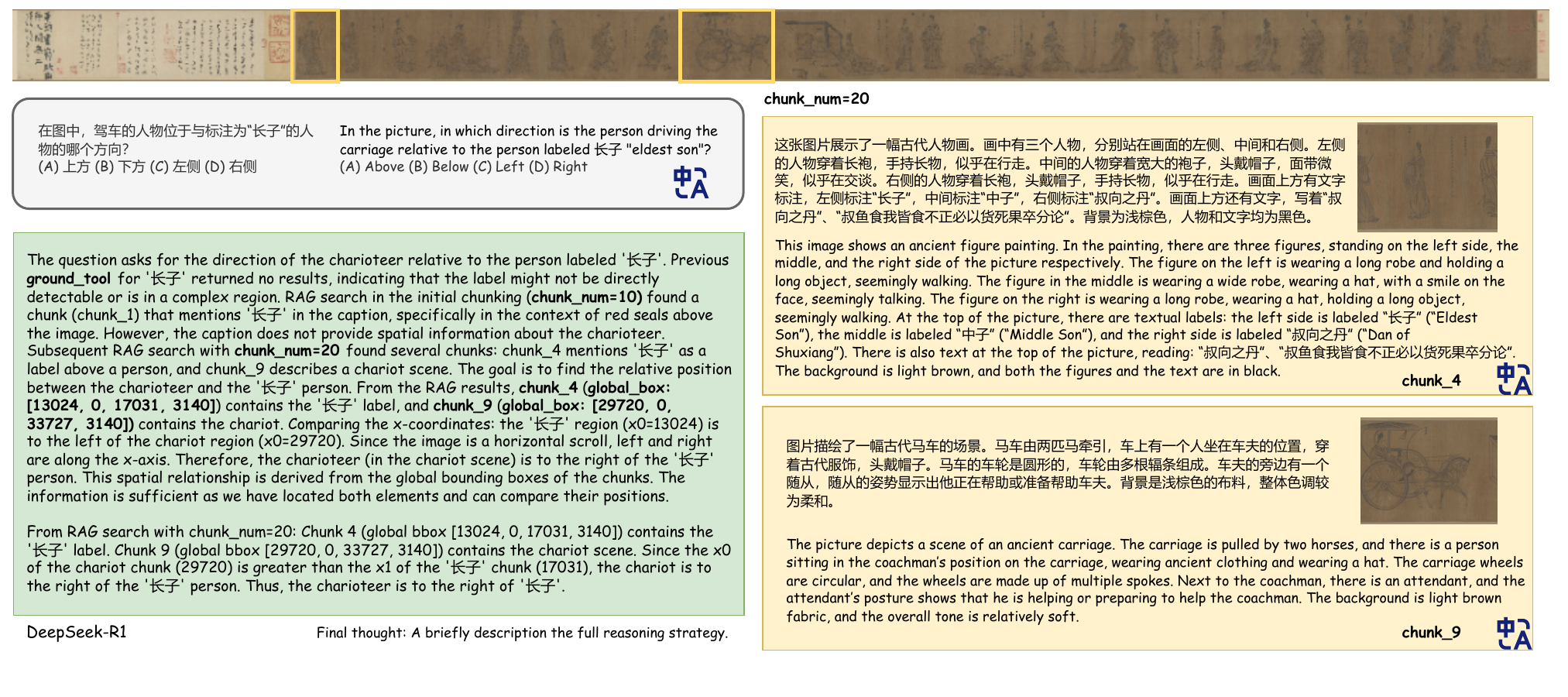}
    \caption{A case from our agent framework under the Humanistic Scenes-Portrait Scrolls subset.}\label{fig:case}
\end{figure*}

\subsection{Tool Usage Analysis}
We analyze the average tool calls across six tools: Semantic Abstraction, Semantic Retrieval, Crop, Enhance, Ground, and VLM. As shown in Figure~\ref{fig:tool_usage}, the left panel illustrates per-tool usage, while the right panel presents the overall average tool calls for each model. 
Models exhibit distinct levels of tool reliance. doubao-seed-1-6-thinking shows the highest overall average calls (around 2.0), followed by DeepSeek-R1 and Qwen3-235B-A22B-Instruct, indicating stronger dependence on external tools for multimodal reasoning. In contrast, models such as gpt4o, claude-sonnet-4, and especially gpt-4.1 demonstrate lower tool usage (below 1.0), suggesting a greater degree of internalized visual reasoning ability. Notably, gemini-2.5-flash-thinking, with few tool calls, performs poorly, as shown in Table~\ref{acc}.
These results suggest a clear stratification in model behavior. And the dominance of the semantic tools further indicates that semantic abstraction and retrieval form the core of the ultra–high-resolution reasoning.


\subsection{Qualitative Analysis}

Figure~\ref{fig:case} illustrates how our Semantic Abstraction framework enables flexible control over reasoning granularity through agent-guided chunk selection. The green box shows the agent’s final summarized reasoning, which integrates evidence drawn from different abstraction levels. On the right, the yellow boxes display the captions of the two chunks referenced by the agent. Notably, by increasing the chunk budget from a coarse setting (e.g., {chunk\_num=10}) to a finer one (e.g., {chunk\_num=20}), the agent obtains more localized and semantically precise chunks, allowing it to refine its interpretation and arrive at a more accurate judgment. This case demonstrates that Semantic Abstraction provides an effective mechanism for adjusting the semantic granularity of visual information, enabling the agent to dynamically adapt its reasoning process to task complexity.

\section{Conclusion}

We introduced \textbf{UR-Bench}, a benchmark targeting multi-hop reasoning over ultra–high-resolution images, and showed that existing MLLMs face substantial challenges under such extreme visual conditions. To address these limitations, we proposed an \textbf{agent-based framework} that leverages Semantic Abstraction and Retrieval to efficiently process and reason over large-scale visual content. Experiments validate the difficulty of UR-Bench and the effectiveness of our framework. 
We hope this work inspires research on tool-augmented high-resolution multimodal understanding.
{
    \small
    \bibliographystyle{ieeenat_fullname}
    \bibliography{main}

@String(AAAI = {AAAI})

@article{shao2021sunet,
  title={SUNet: Change detection for heterogeneous remote sensing images from satellite and UAV using a dual-channel fully convolution network},
  author={Shao, Ruizhe and Du, Chun and Chen, Hao and Li, Jun},
  journal={Remote Sensing},
  volume={13},
  number={18},
  pages={3750},
  year={2021},
  publisher={MDPI}
}

@inproceedings{HRbench,
  title={Divide, conquer and combine: A training-free framework for high-resolution image perception in multimodal large language models},
  author={Wang, Wenbin and Ding, Liang and Zeng, Minyan and Zhou, Xiabin and Shen, Li and Luo, Yong and Yu, Wei and Tao, Dacheng},
  booktitle={Proceedings of the AAAI Conference on Artificial Intelligence},
  volume={39},
  number={8},
  pages={7907--7915},
  year={2025}
}

@inproceedings{vstar,
  title={V*: Guided visual search as a core mechanism in multimodal llms},
  author={Wu, Penghao and Xie, Saining},
  booktitle={Proceedings of the IEEE/CVF Conference on Computer Vision and Pattern Recognition},
  pages={13084--13094},
  year={2024}
}

@article{bai2025qwen2,
  title={Qwen2. 5-vl technical report},
  author={Bai, Shuai and Chen, Keqin and Liu, Xuejing and Wang, Jialin and Ge, Wenbin and Song, Sibo and Dang, Kai and Wang, Peng and Wang, Shijie and Tang, Jun and others},
  journal={arXiv preprint arXiv:2502.13923},
  year={2025}
}

@inproceedings{liu2024grounding,
  title={Grounding dino: Marrying dino with grounded pre-training for open-set object detection},
  author={Liu, Shilong and Zeng, Zhaoyang and Ren, Tianhe and Li, Feng and Zhang, Hao and Yang, Jie and Jiang, Qing and Li, Chunyuan and Yang, Jianwei and Su, Hang and others},
  booktitle={European conference on computer vision},
  pages={38--55},
  year={2024},
  organization={Springer}
}

@article{chen2024bge,
  title={Bge m3-embedding: Multi-lingual, multi-functionality, multi-granularity text embeddings through self-knowledge distillation},
  author={Chen, Jianlv and Xiao, Shitao and Zhang, Peitian and Luo, Kun and Lian, Defu and Liu, Zheng},
  journal={arXiv preprint arXiv:2402.03216},
  year={2024}
}

@article{su2025thinking,
  title={Thinking with images for multimodal reasoning: Foundations, methods, and future frontiers},
  author={Su, Zhaochen and Xia, Peng and Guo, Hangyu and Liu, Zhenhua and Ma, Yan and Qu, Xiaoye and Liu, Jiaqi and Li, Yanshu and Zeng, Kaide and Yang, Zhengyuan and others},
  journal={arXiv preprint arXiv:2506.23918},
  year={2025}
}

@article{li2025imagine,
  title={Imagine while reasoning in space: Multimodal visualization-of-thought},
  author={Li, Chengzu and Wu, Wenshan and Zhang, Huanyu and Xia, Yan and Mao, Shaoguang and Dong, Li and Vuli{\'c}, Ivan and Wei, Furu},
  journal={arXiv preprint arXiv:2501.07542},
  year={2025}
}

@article{su2025openthinkimg,
  title={Openthinkimg: Learning to think with images via visual tool reinforcement learning},
  author={Su, Zhaochen and Li, Linjie and Song, Mingyang and Hao, Yunzhuo and Yang, Zhengyuan and Zhang, Jun and Chen, Guanjie and Gu, Jiawei and Li, Juntao and Qu, Xiaoye and others},
  journal={arXiv preprint arXiv:2505.08617},
  year={2025}
}

@article{ma2024taco,
  title={Taco: Learning multi-modal action models with synthetic chains-of-thought-and-action},
  author={Ma, Zixian and Zhang, Jianguo and Liu, Zhiwei and Zhang, Jieyu and Tan, Juntao and Shu, Manli and Niebles, Juan Carlos and Heinecke, Shelby and Wang, Huan and Xiong, Caiming and others},
  journal={arXiv preprint arXiv:2412.05479},
  year={2024}
}

@inproceedings{gupta2023visual,
  title={Visual programming: Compositional visual reasoning without training},
  author={Gupta, Tanmay and Kembhavi, Aniruddha},
  booktitle={Proceedings of the IEEE/CVF conference on computer vision and pattern recognition},
  pages={14953--14962},
  year={2023}
}

@article{comanici2025gemini,
  title={Gemini 2.5: Pushing the frontier with advanced reasoning, multimodality, long context, and next generation agentic capabilities},
  author={Comanici, Gheorghe and Bieber, Eric and Schaekermann, Mike and Pasupat, Ice and Sachdeva, Noveen and Dhillon, Inderjit and Blistein, Marcel and Ram, Ori and Zhang, Dan and Rosen, Evan and others},
  journal={arXiv preprint arXiv:2507.06261},
  year={2025}
}

@article{xu2025qwen3,
  title={Qwen3-omni technical report},
  author={Xu, Jin and Guo, Zhifang and Hu, Hangrui and Chu, Yunfei and Wang, Xiong and He, Jinzheng and Wang, Yuxuan and Shi, Xian and He, Ting and Zhu, Xinfa and others},
  journal={arXiv preprint arXiv:2509.17765},
  year={2025}
}

@article{wang2025internvl3,
  title={Internvl3. 5: Advancing open-source multimodal models in versatility, reasoning, and efficiency},
  author={Wang, Weiyun and Gao, Zhangwei and Gu, Lixin and Pu, Hengjun and Cui, Long and Wei, Xingguang and Liu, Zhaoyang and Jing, Linglin and Ye, Shenglong and Shao, Jie and others},
  journal={arXiv preprint arXiv:2508.18265},
  year={2025}
}

@article{yu2024visrag,
  title={Visrag: Vision-based retrieval-augmented generation on multi-modality documents},
  author={Yu, Shi and Tang, Chaoyue and Xu, Bokai and Cui, Junbo and Ran, Junhao and Yan, Yukun and Liu, Zhenghao and Wang, Shuo and Han, Xu and Liu, Zhiyuan and others},
  journal={arXiv preprint arXiv:2410.10594},
  year={2024}
}

@inproceedings{OCRVQA,
  title={Ocr-vqa: Visual question answering by reading text in images},
  author={Mishra, Anand and Shekhar, Shashank and Singh, Ajeet Kumar and Chakraborty, Anirban},
  booktitle={2019 international conference on document analysis and recognition (ICDAR)},
  pages={947--952},
  year={2019},
  organization={IEEE}
}

@inproceedings{STVQA,
  title={Latr: Layout-aware transformer for scene-text vqa},
  author={Biten, Ali Furkan and Litman, Ron and Xie, Yusheng and Appalaraju, Srikar and Manmatha, R},
  booktitle={Proceedings of the IEEE/CVF conference on computer vision and pattern recognition},
  pages={16548--16558},
  year={2022}
}

@inproceedings{DocVQA,
  title={Docvqa: A dataset for vqa on document images},
  author={Mathew, Minesh and Karatzas, Dimosthenis and Jawahar, CV},
  booktitle={Proceedings of the IEEE/CVF winter conference on applications of computer vision},
  pages={2200--2209},
  year={2021}
}

@inproceedings{TextVQA,
  title={Towards vqa models that can read},
  author={Singh, Amanpreet and Natarajan, Vivek and Shah, Meet and Jiang, Yu and Chen, Xinlei and Batra, Dhruv and Parikh, Devi and Rohrbach, Marcus},
  booktitle={Proceedings of the IEEE/CVF conference on computer vision and pattern recognition},
  pages={8317--8326},
  year={2019}
}

@inproceedings{ChartQA,
  title={Chartqa: A benchmark for question answering about charts with visual and logical reasoning},
  author={Masry, Ahmed and Do, Xuan Long and Tan, Jia Qing and Joty, Shafiq and Hoque, Enamul},
  booktitle={Findings of the association for computational linguistics: ACL 2022},
  pages={2263--2279},
  year={2022}
}

@article{wang2024charxiv,
  title={Charxiv: Charting gaps in realistic chart understanding in multimodal llms},
  author={Wang, Zirui and Xia, Mengzhou and He, Luxi and Chen, Howard and Liu, Yitao and Zhu, Richard and Liang, Kaiqu and Wu, Xindi and Liu, Haotian and Malladi, Sadhika and others},
  journal={Advances in Neural Information Processing Systems},
  volume={37},
  pages={113569--113697},
  year={2024}
}

@techreport{anthropic2025claude37,
  title        = {Claude 3.7 Sonnet System Card},
  author       = {Anthropic},
  year         = {2025},
  month        = feb,
  url          = {https://www.anthropic.com/claude-3-7-sonnet-system-card},
  institution  = {Anthropic},
}

@techreport{openai2025o3,
  title        = {OpenAI o3 System Card},
  author       = {OpenAI},
  year         = {2025},
  month        = apr,
  url          = {https://openai.com/research/o3-system-card},
  institution  = {OpenAI},
  note         = {Accessed: 2025-09-23}
}

@article{hurst2024gpt,
  title={Gpt-4o system card},
  author={Hurst, Aaron and Lerer, Adam and Goucher, Adam P and Perelman, Adam and Ramesh, Aditya and Clark, Aidan and Ostrow, AJ and Welihinda, Akila and Hayes, Alan and Radford, Alec and others},
  journal={arXiv preprint arXiv:2410.21276},
  year={2024}
}

@article{guo2025deepseek,
  title={Deepseek-r1: Incentivizing reasoning capability in llms via reinforcement learning},
  author={Guo, Daya and Yang, Dejian and Zhang, Haowei and Song, Junxiao and Zhang, Ruoyu and Xu, Runxin and Zhu, Qihao and Ma, Shirong and Wang, Peiyi and Bi, Xiao and others},
  journal={arXiv preprint arXiv:2501.12948},
  year={2025}
}
}


\end{document}